\pdfoutput=1
\documentclass[11pt,a4paper]{article}
\usepackage{times}
\usepackage{acl2021}
\usepackage{graphicx}
\usepackage{latexsym}

\usepackage{microtype}

\aclfinalcopy 
\def\aclpaperid{98} 

\title{Dialogue State Tracking with Multi-Level Fusion of Predicted Dialogue States and Conversations}

\author{Jingyao Zhou$^{1,2}$ \And Haipang Wu$^{1,2}$ \And Zehao Lin$^{3}$ \\
$^{1}$Hithink RoyalFlush Information Network Co., Ltd\\
$^{2}$Hithink RoyalFlush AI Research Institute\\
$^{3}$College of Computer Science and Technology, Zhejiang University\\
\texttt{$\{$zhoujingyao,wuhaipang$\}$@myhexin.com}\\
\texttt{$\{$georgenlin,guodun.li,zhangyin98$\}$@zju.edu.cn}
\And Guodun Li$^{3}$ \And Yin Zhang$^{3}\thanks{\ \ Corresponding Author}$
 }

\date{}

\begin{document}
\maketitle
\begin{abstract}
Most recently proposed approaches in dialogue state tracking (DST) leverage the context and the last dialogue states to track current dialogue states, which are often slot-value pairs. Although the context contains the complete dialogue information, the information is usually indirect and even requires reasoning to obtain. The information in the lastly predicted dialogue states is direct, but when there is a prediction error, the dialogue information from this source will be incomplete or erroneous. In this paper, we propose the Dialogue State Tracking with Multi-Level \textbf{F}usion of \textbf{P}redicted  \textbf{D}ialogue \textbf{S}tates and \textbf{C}onversations network (FPDSC). This model extracts information of each dialogue turn by modeling interactions among each turn utterance, the corresponding last dialogue states, and dialogue slots. Then the representation of each dialogue turn is aggregated by a hierarchical structure to form the passage information, which is utilized in the current turn of DST. Experimental results validate the effectiveness of the fusion network with 55.03$\%$ and 59.07$\%$ joint accuracy on MultiWOZ 2.0 and MultiWOZ 2.1 datasets, which reaches the state-of-the-art performance. Furthermore, we conduct the deleted-value and related-slot experiments on MultiWOZ 2.1 to evaluate our model.
\end{abstract}

\section{Introduction}

Dialogue State Tracking (DST) is utilized by the dialogue system to track dialogue-related constraints and user's requests in the dialogue context. Traditional dialogue state tracking models combine semantics extracted by language understanding modules to estimate the current dialogue states \citep{williams2007partially,thomson2010bayesian,wang2013simple,williams2014web}, or to jointly learn speech understanding \citep{henderson2014word,zilka2015incremental,wen2016network}.
They rely on hand-crafted features and complex domain-specific lexicons, which are vulnerable to linguistic variations and difficult to scale. Recently proposed approaches attempt to automatically learn features from the dialogue context and the previous dialogue states. Most of them utilize only the context \citep{shan2020contextual} or encode the concatenation of context and dialogue states \citep{hosseini2020simple} or utilize a simple attention mechanism to merge the information from the above two sources \citep{ouyang2020dialogue}. These methods do not fully exploit the nature of the information in the context and the predicted dialogue states. The information nature of the context is complete and may be indirect. While the nature of the predicted dialogue states is direct and may be incomplete or erroneous.

\begin{table}
\small
\begin{tabular}{l}
\hline
$U_{1}$: I need a place to dine in the centre.\\
\textbf{State:} restaurant-area=centre\\
\hline
$S_{2}$:I recommend the rice house. Would you like me \\to reserve a table?\\
$U_{2}$:Yes, please book me a table for 9.\\
\textbf{State:} restaurant-area=centre; restaurant-book people=9; \\\textit{restaurant-name=rice house}\\
\hline
$S_{3}$:Unfortunately, I could not book the rice house for that\\ amount of people.\\ 
$U_{3}$: please find another restaurant for that amount of\\ people at that time. \\
\textbf{State:} restaurant-area=centre; 
restaurant-book people=9\\\textit{restaurant-name=none}\\
\hline
$S_{4}$: how about tang restaurant ?\\
$U_{4}$: Yes, please make me a reservation. I also need \\a taxi.\\
\textbf{State:} restaurant-area=centre; restaurant-book people=9\\
\textit{restaurant-name=tang};\\
\hline
$S_{5}$: What is your destination ?\\
$U_{5}$: To the restaurant.\\
\textbf{State:} restaurant-area=centre; restaurant-book people=9\\
\textit{restaurant-name=tang}; \textit{taxi-destination=tang}\\
\hline
\end{tabular}
\caption{\label{dialogue_example} An example of dialogue contains (1) the \emph{deleted-value problem} at the $3^{rd}$ turn, which changes restaurant-name from rice house to none, and (2) the \emph{related-slot phenomenon} at the $5^{th}$ turn, which carries over the value from restaurant-name to taxi-destination.}
\end{table}

Our FPDSC model exploits the interaction among the turn utterance, the corresponding last dialogue states, and dialogue slots at each turn. A fusion gate (the turn-level fusion gate) is trained to balance the keep-proportion of the slot-related information from the turn utterance and the corresponding last dialogue states at each turn. Then it applies a hierarchical structure to keep the complete information of all dialogue turns. On top of the model, we employ another fusion gate (the passage-level fusion gate) to strengthen the impact of the last dialogue states. \citet{ouyang2020dialogue} shows that such strengthening is vital to solve the related-slot problem. The problem is explained in Table \ref{dialogue_example}. To eliminate the negative impact of the error in the predicted dialogue states, we train our models in two phases. In the teacher-forcing phase, previous dialogue states are all true labels. While in the uniform scheduled sampling phase \citep{bengio2015scheduled}, previous dialogue states are half predicted dialogue states and half true labels. Training with such natural data noise from the error in the predicted dialogue states helps improve the model's robustness.

 For ablation studies, we test the following variants of FPDSC: base model (without turn/passage-level fusion gates), turn-level model (with only turn-level fusion gate), passage-level model (with only passage-level fusion gate) and dual-level model (with both turn/passage-level fusion gates). We also conduct the experiment for the deleted-value problem, which is explained in Table \ref{dialogue_example}, and the related-slot problem. Besides, we design two comparative networks to validate the effectiveness of the turn-level fusion gate and the whole previous dialogue states. One comparative network employs only the attention mechanism to merge information from the turn utterance, the corresponding last dialogue states, and dialogue slots at each turn. Another comparative network utilize only the last previous dialogue states in the turn-level fusion gate. Our model shows strong performance on MultiWOZ 2.0 \citep{budzianowski2018multiwoz} and MultiWOZ 2.1 \citep{eric2019multiwoz} datasets. Our main contributions are as follows: 

\begin{itemize}
\item We propose a novel model, which utilizes multi-level fusion gates and the attention mechanism to extract the slot-related information from the conversation and previous dialogue states. The experimental results of two comparative networks validate the effectiveness of the turn-level fusion gate to merge information and the importance of the whole previous dialogue states to improve DST performance.

\item Both turn/passage-level fusion between the context and the last dialogue states helps at improving the model's inference ability. The passage-level fusion gate on the top of the model is more efficient than the turn-level fusion gate on the root for slot correlation problem. While the turn-level fusion gate is sensitive to signal tokens in the utterance, which helps improve the general DST performance. 

\item Experimental results on the deleted-value and the related-slot experiment shows the ability of the structure to retrieve information. Besides, our models reach state-of-the-art performance on MultiWOZ 2.0/2.1 datasets. 
\end{itemize}

\section{Related Work}
\begin{figure}[t]
\includegraphics[width=4.8cm]{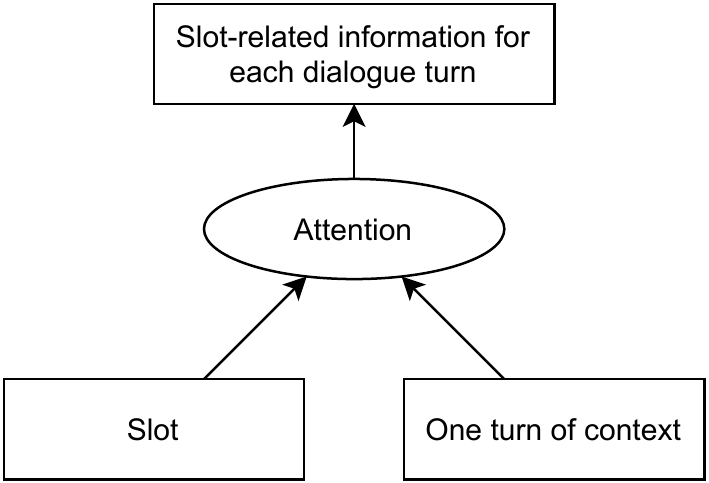}
\centering
\caption{Part of CHAN and FPDSC (base/passage-level)}
\label{chan-pdsc-1}
\end{figure}

\begin{figure}[t]
\centering
\includegraphics[width=6.8cm]{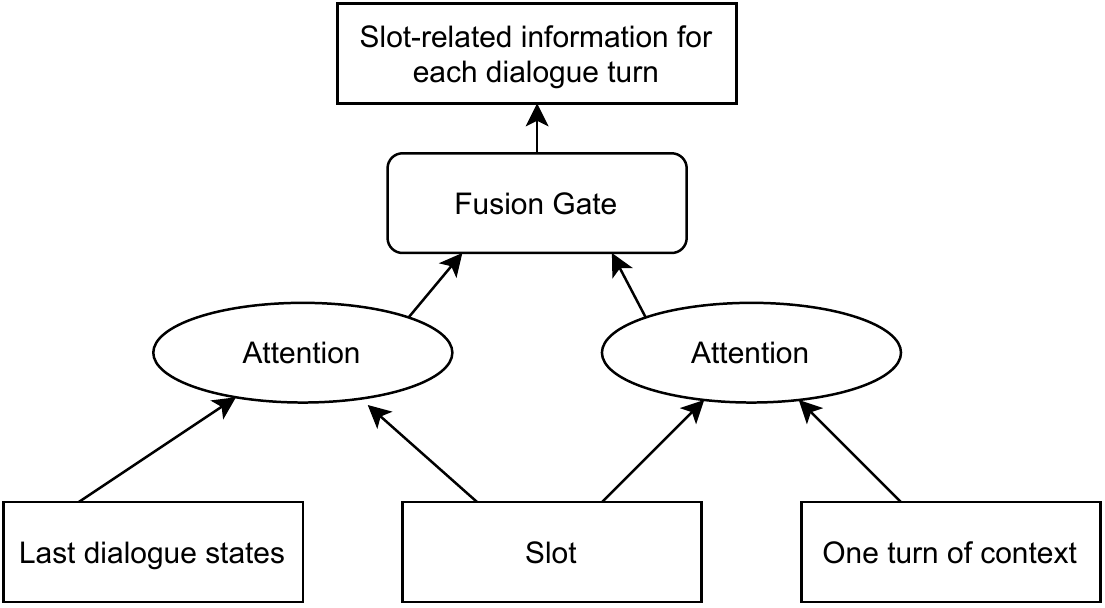}
\caption{Part of FPDSC (turn/dual-level)}
\label{chan-pdsc-2}
\end{figure}

Recently proposed methods show promising progress in the challenge of DST. CHAN \citep{shan2020contextual} employs a contextual hierarchical attention network, which extracts slot attention based representation from the context in both token- and utterance-level. Benefiting from the hierarchical structure, CHAN can effectively keep the whole dialogue contextual information. Although CHAN achieves the new state-of-the-art performance on MultiWoz 2.0/2.1 datasets, it ignores the information from the predicted dialogue states. Figures \ref{chan-pdsc-1} and \ref{chan-pdsc-2}  show the difference between CHAN and FPDSC in the extraction of the slot-related information in one dialogue turn.

In the work of \citet{ouyang2020dialogue}, the problem of slot correlations across different domains is defined as related-slot problem. DST-SC \citep{ouyang2020dialogue} model is proposed. In the approach, the last dialogue states are vital to solve the related-slot problem. The method merges slot-utterance attention result and the last dialogue states with an attention mechanism. However, the general performance of DST-SC is worse than CHAN. 

SOM-DST \citep{kim2019efficient} and CSFN-DST \citet{zhu2020efficient} utilize part of the context and the last dialogue states as information sources. The two methods are based on the assumption of Markov property in dialogues. They regard the last dialogue states as a compact representation of the whole dialogue history. Once a false prediction of a slot exists and the slot-related context is dropped, the dialogue states will keep the error. 

\section{Model}
\begin{figure*}[t]
\centering
\resizebox{1\linewidth}{!}{
\includegraphics[width=1\textwidth]{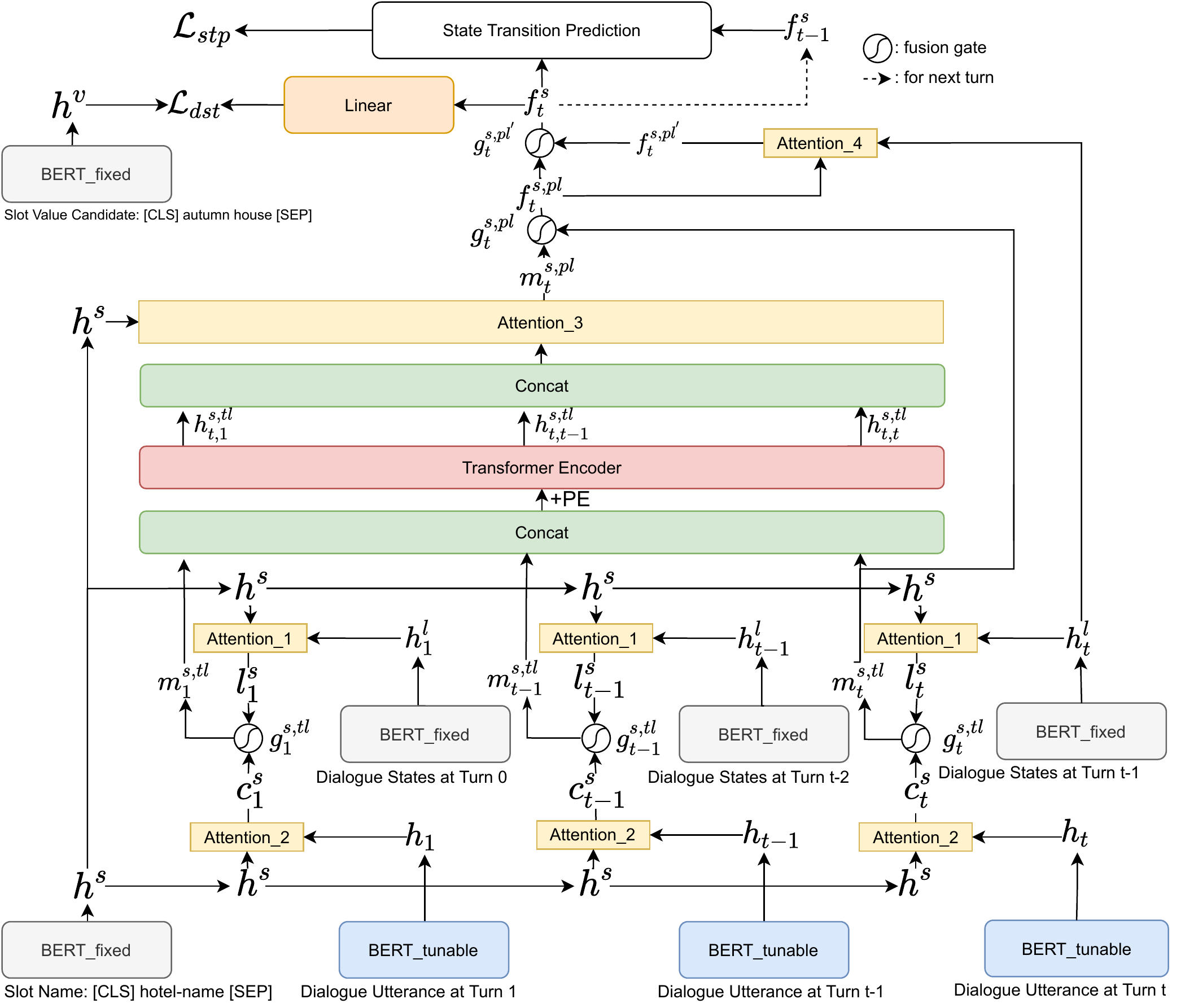}}
\caption{The structure of \textbf{dual-level} FPDSC. The dialogue utterance at the $t$-th turn is [CLS]{$R_t$}[SEP]{$U_t$}[SEP]. The dialogue state is a list of slot-value pairs ([CLS]Slot$_1$[SEP]Value$_1$[SEP],$\dots$,[CLS]Slot$_n$[SEP]Value$_n$[SEP]). All slot values are none in the initial dialogue states. The \textbf{turn-level} approach is without Attention$\_4$ and the passage-level fusion gate ($g_t^{s,pl^{'}}$ is the output weight of the gate). The \textbf{passage-level} approach is without Attention$\_1$ and the turn-level fusion gate ($\{g_1^{s,tl},\cdots, g_{t-1}^{s,tl}, g_{t}^{s,tl}\}$ are the output weights of the gate). The \textbf{base} approach is without Attention$\_1/4$ and turn/passage-level fusion gate. The \textbf{base} approach has the same structure as CHAN with different early stop mechanism.}
\label{fig1}
\end{figure*}
Figure \ref{fig1} shows the overall structure of FPDSC (dual-level). The followings are important notations for our model.

\noindent
\textbf{Inputs:} The context {$D = \{(U_{1}, R_{1}), \dots, (U_{t}, R_{t})\}$} where $U_{t}$ and $R_{t}$ represent utterance for user and system at the $t$-th dialogue turn; The previous dialogue states $B = \{B_{1}, \dots ,B_{t-1}\}$ where $B_{t} = \{(s, v_{t}), s \in \mathcal{S}\}$, $\mathcal{S}$ is slot set, $s$ is one of the slot names, $v_{t}$ is the corresponding slot value at the $t$-th turn; $\mathcal{V} = \{{V_{s}, s \in \mathcal{S}}\}$ are slot value candidates of all slots .

\noindent
\textbf{Turn-level Information:} The slot-related information for each dialogue turn in Figure \ref{chan-pdsc-2} is the turn-level information.
 In Figure \ref{fig1}, the turn-level information is denoted as $\{m_1^{s,tl}, \dots, m_{t-1}^{s,tl}, m_{t}^{s,tl}\}$, which is the fusion (the turn-level fusion gate) result of the slot-utterance attention results $\{c_l^{s}, \dots, c_{t-1}^{s}, c_{t}^{s}\}$ and the slot-dialogue-states attention results $\{l_l^{s}, \dots, l_{t-1}^{s}, l_{t}^{s}\}$. The weights $\{g_1^{s,tl},\dots,g_{t-1}^{s,tl}, g_{t}^{s,tl} \}$ are from the same fusion gate, which is utilized to allocate the keep-proportion from the conversations and previous dialogue states. The turn-level information of a slot is fed to a transformer encoder to form the \textbf{mutual interaction information}  $\{h_{t,1}^{s,tl},\cdots, h_{t,t-1}^{s,tl}, h_{t,t}^{s,tl}\}$. 

\noindent
\textbf{Passage-level Information:} The attention $\{\mathrm{Attention}\_{3}\}$ result of the \textbf{mutual interaction information} and a slot is the passage-level information $\{m_t^{s,pl}\}$ of a slot. 

\noindent
\textbf{Core Feature:} The weight $\{g_t^{s,pl}\}$ are applied to balance the turn-level information of the current dialogue turn $\{m_t^{s,tl}\}$ and the passage-level information $\{m_t^{s,pl}\}$ of a slot. We employ the attention $\{\mathrm{Attention}\_{4}\}$ mechanism between the turn/passage-level balanced information $\{f_t^{s,pl}\}$ and the last dialogue states $\{h_t^{l}\}$ to strengthen the impact of the last dialogue states. Another weight $\{g_t^{s,pl^{'}}\}$ (from the passage-level fusion gate) merge the turn/passage-level balanced information $\{f_t^{s,pl}\}$ and the strengthened information $\{f_t^{s,pl^{'}}\}$ to form the core feature $\{f_t^{s}\}$, which is utilized in the downstream tasks.

\subsection{BERT-Base Encoder}
Due to pre-trained models' (e.g., BERT) strong language understanding capabilities \citep{mehri2020dialoglue}, we use the fixed-parameter BERT-Base encoder ($\mathrm{BERT}_{fixed}$) to extract the representation of slot names, slot values and the previous dialogue states. Three parts share the same parameters from HuggingFace \footnote{https://huggingface.co/}. We also apply a tunable BERT-Base encoder ($\mathrm{BERT}_{tunable}$) to learn the informal and noisy utterances distribution \citep{zhang2019dialogpt} in the dialogue context. The two BERT-Base Encoders are input layers of the model. [CLS] and [SEP] represent the beginning and the end of a text sequence. We use the output at [CLS] to represent the whole text for $\mathrm{BERT}_{fixed}$. A slot-value pair in the last dialogue states at the $t$-th turn is denoted as:
\begin{equation}
h_{t}^{ls} = \mathrm{BERT}_{fixed}([s;v_{t-1}]) \label{last dialogue state unit}
\end{equation}
where $h_{t}^{ls}$ is the $s$ slot-related representation of last dialogue state at the dialogue $t$-th turn. Thus the full representation of the last dialogue states at the $t$-th turn is as follows:
\begin{equation}
h_{t}^{l} = h_{t}^{ls_1}\oplus\cdots h_{t}^{ls_k}\cdots\oplus h_{t}^{ls_{n}} \quad ls_k\in\mathcal{S}
\end{equation}
$\oplus$ means concatenation. The entire history of the dialogue states is $H_t^{p} = \{h_{1}^{l},\cdots, h_{t-1}^{l},h_{t}^{l} \}$. The representations of slot $s$ and its corresponding value $v_t$ are as follows: 
\begin{equation}
h^{s} = \mathrm{BERT}_{fixed}(s)
\end{equation}
\begin{equation}
h^{v} = \mathrm{BERT}_{fixed}(v_t)
\end{equation}
$\mathrm{BERT}_{tunable}$ extracts utterances distribution of user $U_t = \{w_1^u, \cdots, w_l^u\}$ and system $R_t = \{w_1^r, \cdots, w_{l}^r\}$ at the $t$-th turn, which are marked as:
\begin{equation}
h_{t} = \mathrm{BERT}_{tunable}([R_t;U_t])
\end{equation}
The dialogue context until $t$-th turn is $H_t = \{h_1, \cdots, h_{t-1}, h_{t}\}$.

\subsection{MultiHead-Attention Unit}
We utilize MultiHead-Attention \citep{vaswani2017attention} here to get the slot-related information from the turn utterance and the corresponding last dialogue states. The representations at the $t$-th turn are as follows:
\begin{equation}\label{attention_1}
l_{t}^{s}= \mathrm{Attention\_1}(h^{s},h_t^{l},h_t^{l})
\end{equation}
\begin{equation}\label{attention_2}
c_{t}^{s} = \mathrm{Attention\_2}(h^s,h_t,h_t)
\end{equation}
Another attention unit is applied to get the passage-level information of a slot from the mutual interaction information  
$H_t^{s, tl} = \{h_{t,1}^{s, tl},\cdots, h_{t,t-1}^{s, tl}, h_{t,t}^{s, tl}\}$, which is described in section \ref{Transformer Encoder}.
\begin{equation}\label{attention_3}
m_{t}^{s,pl}=\mathrm{Attention\_3}(h^s, H_t^{s, tl}, H_t^{s, tl})
\end{equation}
We apply an attention unit to connect the representation of the merged turn/passage-level balanced information $f_t^{s,pl}$ and the last dialogue states to enhance the impact of the last dialogue states.
\begin{equation}\label{attention_4}
f_t^{s, pl^{'}} = \mathrm{Attention\_4}(f_{t}^{s,pl} ,h_t^l,h_t^l)
\end{equation}
$f_t^{s, pl^{'}}$ is the enhanced result. All attention units above do not share parameters.

\subsection{Transformer Encoder} \label{Transformer Encoder}
The complete turn-level merged information $M_t^{s,tl} = \{m_{1}^{s,tl},\cdots, m_{t-1}^{s,tl},m_{t}^{s,tl}\}$ has no dialogue sequence information. Besides, each turn representation does not fully share information. Thus we apply a transformer encoder \citep{vaswani2017attention}. 
\begin{equation}
H_t^{s, tl} = \mathrm{TransformerEncoder}(M_t^{s,tl})
\end{equation}
where $H_t^{s, tl} = \{h_{t,1}^{s, tl},\cdots, h_{t,t-1}^{s, tl}, h_{t,t}^{s, tl}\}$ means the mutual interaction information. $h_{t,1}^{s,tl}$ means the $s$ slot-related representation of the $1^{st}$ dialogue turn after turn interaction, when the dialogue comes to the $t$-th turn. The transformer encoder utilizes positional encoding to record the position information and self-attention to get interacted information in each dialogue turn.  

\subsection{Fusion Gate}
Fusion gate is applied to merge the information as follows:
\begin{equation}
g_{t}^{s,tl} = \sigma (W_{tl}\odot[c_t^s;l_t^s])
\end{equation}
\begin{equation}
m_{t}^{s,tl} = (1 - g_{t}^{s,tl})\otimes c_t^s + g_{t}^{s,tl} \otimes l_t^s
\end{equation}
$\odot$ and $\otimes$ mean the matrix product and point-wise product. $\sigma$ is the sigmoid function. $g_{t}^{s,tl}$ is the output weight of the fusion gate to keep the information from the last dialogue state. $M_t^{s,tl} = \{m_{1}^{s,tl},\cdots, m_{t-1}^{s,tl},m_{t}^{s,tl}\}$ is the turn-level information;
\begin{equation}
g_t^{s, pl} = \sigma(W_{pl}\odot[m_t^{s,tl};m_t^{s,pl}])
\end{equation}
\begin{equation}
f_{t}^{s,pl} = (1-g_t^{s, pl})\otimes m_t^{s,pl} + g_t^{s, pl}\otimes m_t^{s,tl}
\end{equation}
$g_t^{s, pl}$ is the weight to balance the turn-level merged information $m_{t}^{s,tl}$ and the passage-level extracted information $m_t^{s,pl}$;
\begin{equation}
g_t^{s, pl^{'}} = \sigma(W_{pl^{'}}\odot[f_{t}^{s,pl};f_{t}^{s,pl^{'}}])
\end{equation}
\begin{equation}
f_{t}^{s} = (1-g_t^{s, pl^{'}})\otimes f_{t}^{s,pl} + g_t^{s, pl}\otimes f_{t}^{s,pl^{'}}
\end{equation}
$g_t^{s, pl^{'}}$ is the weight to balance the merged turn/passage-level balanced information $f_{t}^{s,pl}$ and the enhanced result $f_{t}^{s,pl^{'}}$ from equation \ref{attention_4}. $f_t^{s}$ is  $s$ slot-related core feature from context and the entire history of dialogue states.

\subsection{Loss Function}
Here we follow \citet{shan2020contextual} to calculate the probability distribution of value $v_t$ and predict whether the slot $s$ should be updated or kept compared to the last dialogue states. Thus our loss functions are as follows:
\begin{equation}
o_{t}^{s} = \mathrm{LayerNorm}(\mathrm{Linear}(\mathrm{Dropout}(f_t^{s})))
\end{equation}
\begin{equation}
p(v_{t}|U_{\le t}, R_{\le t}, s) =\frac{\mathrm{exp}(-\|o^s_t-h^v\|_{2})}{\sum\limits_{v^{'} \in \mathcal{V}_{s}}\mathrm{exp}(-\|o^s_t-h^{v^{'}}\|_{2})}
\end{equation}
\begin{equation}
\mathcal{L}_{dst} = \sum\limits_{s\in\mathcal{S}}\sum\limits_{t=1}^{T}-\mathrm{log}(p(\widehat{v}_{t}|U_{\le t}, R_{\le t}, s))
\end{equation}
$\mathcal{L}_{dst}$ is the distance loss for true value $\widehat{v}$ of slot $s$;
\begin{equation}
c_t^{s, stp} = \mathrm{tanh}(W_c\odot f_t^{s})
\end{equation}
\begin{equation}
p_t^{s,stp} = \sigma(W_p\odot[c_t^{s, stp};c_{t-1}^{s, stp}])
\end{equation}
\begin{equation}
\mathcal{L}_{stp} = \sum\limits_{s\in\mathcal{S}}\sum\limits_{t=1}^{T}-y_{t}^{s,stp}\cdot \mathrm{log}({p_{t}^{s,stp}})
\end{equation}
$\mathcal{L}_{stp}$ is the loss function for state transition prediction, which has the value set $\{keep, update\}$. $p_t^{s,stp}$ is update probability for slot $s$ at the $t$-th turn. $y_t^{s,stp}$ is the state transition label with $update$ $y_t^{s,stp}=1$ and $keep$ $y_t^{s,stp}=0$ . We optimize the sum of above loss in the training process:
\begin{equation}
\mathcal{L}_{joint} = \mathcal{L}_{dst}+\mathcal{L}_{stp}
\end{equation}

\begin{table*}
\centering
\resizebox{0.66\textwidth}{!}{
\begin{tabular}{c|c|c}
\hline \textbf{Model} & \textbf{MultiWOZ 2.0} & \textbf{MultiWOZ 2.1} \\
\cline{2-3}
&\textbf{Joint Acc ($\%$)}&\textbf{Joint Acc ($\%$)}\\
\hline
TRADE \citep{wu-etal-2019-transferable} & 48.62 & 46.00 \\
DST-picklist \citep{zhang2019find} & 54.39  & 53.30 \\
TripPy \citep{heck2020trippy} & - & 55.30\\
SimpleTOD \citep{hosseini2020simple} & - & 56.45 \\
CHAN \citep{shan2020contextual} & 52.68 & 58.55 \\
CHAN$^{*}$ \citep{shan2020contextual} & -  & 57.45 \\
\hline 
FPDSC (base) & 51.03 & 54.91\\
FPDSC (passage-level)& 52.31 & 55.86 \\
FPDSC (turn-level)& \textbf{55.03} & 57.88\\
FPDSC (dual-level) & 53.17 & \textbf{59.07}\\
\hline
\end{tabular}
}
\caption{\label{Joint Acc} Joint accuracy on the test sets of MultiWOZ 2.0 and 2.1. CHAN$^{*}$ means performance without adaptive objective fine-tuning, which solves the slot-imbalance problem. CHAN means performance with the full strategy. The overall structure of FPDSC (dual-level) is illustrated in Figure \ref{fig1}.}
\end{table*}

\section{Experiments Setup}
\subsection{Datasets}
We evaluate our model on MultiWOZ 2.0 and MultiWOZ 2.1 datasets. They are multi-domain task-oriented dialogue datasets. MultiWOZ 2.1 identified and fixed many erroneous annotations and user utterances \citep{zang2020multiwoz}.
\subsection{Baseline}
We compare FPDSC with the following approaches:

\noindent
\textbf{TRADE} is composed of an utterance encoder, a slot-gate, and a generator. The approach generates value for every slot using the copy-augmented decoder \citep{wu-etal-2019-transferable}.

\noindent
\textbf{CHAN} employs a contextual hierarchical attention network to enhance the DST. The method applies an adaptive objective to  alleviate the slot imbalance problem \citep{shan2020contextual}.

\noindent
\textbf{DST-picklist} adopts a BERT-style reading comprehension model to jointly handle both categorical and non-categorical slots, matching the value from ontologies \citep{zhang2019find}. 

\noindent
\textbf{TripPy} applies three copy mechanisms to get value span. It regards user input, system inform memory and previous dialogue states as sources \citep{heck2020trippy}.

\noindent
\textbf{SimpleTOD} is an end-to-end approach and regards sub-tasks in the task oriented dialogue task as a sequence prediction problem\citep{hosseini2020simple}.

\subsection{Training Details}
Our code is public \footnote{https://github.com/helloacl/DST-DCPDS}, which is developed based on \textbf{CHAN}'s code \footnote{https://github.com/smartyfh/CHAN-DST}.
In our experiments, we use the Adam optimizer \citep{kingma2014adam}. We use a batch size of 2 and maximal sequence length of 64 for each dialogue turn. The transformer encoder has 6 layers. The multi-head attention units have counts of 4 and hidden sizes of 784. The training process consists of two phases: 1) teacher-forcing training; 2) uniform scheduled sampling \citep{bengio2015scheduled}. The warmup proportion is 0.1 and the peak learning rate is 1e-4. The model is saved according to the best joint accuracy on the validation data. The training process stops with no improvement in 15 continuous epochs. Our training devices are GeForce GTX 1080 Ti and Intel Core i7-6800 CPU@3.40GHZ. The training time of an epoch takes around 0.8 hour in the teacher-forcing phase and 1.6 hours in the uniform scheduled sampling phase with a GPU. 

\section{Results and Analysis}

\begin{table}
\small
\centering
\begin{tabular}{cccc}
\hline
Deleted-Value&&&\\
\hline
Base & Turn & Passage & Dual\\
\hline
$2.84\%$ & $22.87\%$ & $23.98\%$ & $25.22\%$\\
\hline
Related-Slot&&&\\
\hline
Base & Turn & Passage & Dual\\
\hline
$46.63\%$ & $57.85\%$ & $62.23\%$ & $70.85\%$\\
\hline
\end{tabular}
\caption{\label{value_delete_related} Success change rate of the deleted-value and related-slot experiment for FPDSC. Turn, Passage, Dual mean turn-level, passage-level and dual-level FPDSC.}
\end{table}

\subsection{Main Results}
We use the joint accuracy to evaluate the general performance. Table \ref{Joint Acc} shows that our models get 55.03$\%$ and 59.07$\%$ joint accuracy with improvements (0.64$\%$ and 0.52$\%$) over previous best results on MultiWOZ 2.0 and 2.1. All of our approaches get better performance on 2.1 than 2.0.  This is probably because of fewer annotations error in MultiWOZ 2.1. Though Table \ref{value_delete_related} shows that the passage-level variant performs better than the turn-level variant in the deleted-value and the related-slot test, passage-level variant gets worse results in the general test. The small proportion of the above problem in the MultiWOZ dataset and the strong sensitivity of the turn-level fusion gate to signal tokens in the utterance explain the phenomenon.

\subsection{The Comparative Experiment for the Fusion Gate}
\begin{figure}[t]
\includegraphics[width=7cm]{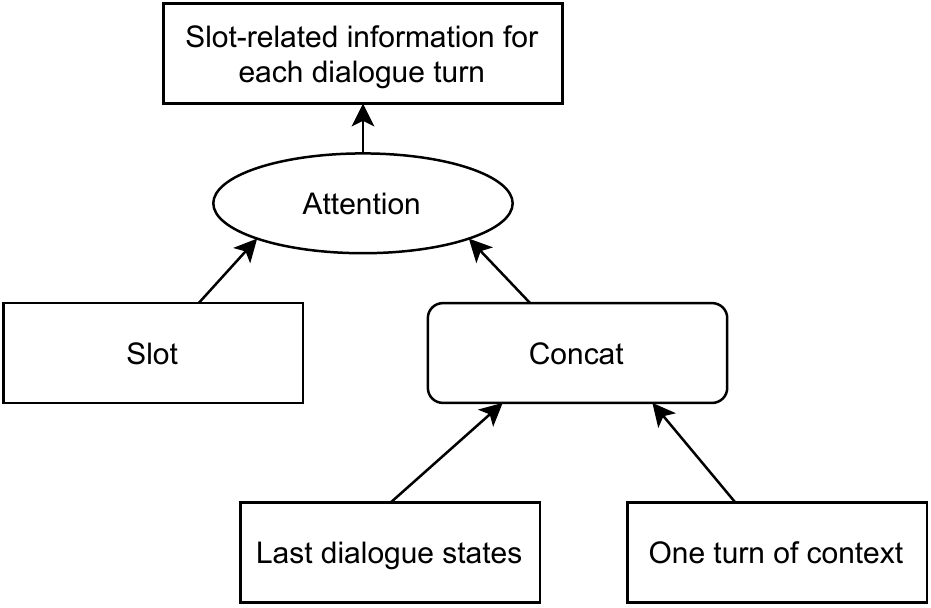}
\caption{Part of the comparative network}
\label{no_gate}
\end{figure}

We design a comparative network to validate the effectiveness of the turn-level fusion gate. Figure \ref{no_gate} shows the part structure of the comparative network (no turn-level fusion gate). The rest of the comparative network is the same as the FPDSC (turn-level). Table \ref{no gate comparative result} shows the performance of the comparative network and the FPDSC (turn-level) on the MultiWOZ 2.1. The result validates the effectiveness of the fusion gate to merge the different information sources. 

\begin{table}
\small
\begin{tabular}{c|c|c|c|c}
\hline
Dataset & no-gate$^{*}$ & no-gate & turn-level$^{*}$ & turn-level \\
\hline
dev  & 46.38 & 52.58 & 56.17 & 61.39 \\
\hline
test &  43.03 & 49.24 & 54.08 & 57.88 \\
\hline
\end{tabular}
\caption{Joint accuracy results of the comparative network (no-gate) and FPDSC (turn-level) on the MultiWOZ 2.1 dataset. $^{*}$ indicates the approach is only trained with teacher-forcing, otherwise is further trained by uniform scheduled sampling after the teacher-forcing phase.}
\label{no gate comparative result}
\end{table}

\subsection{The Comparative Experiment for the Complete Dialogue States }
We design another comparative network to validate the effectiveness of the complete previous dialogue states. As Figure \ref{fig1} shows, $\{m_1^{s,tl}, \dots, m_{t-1}^{s,tl}, m_{t}^{s,tl}\}$ are fed to the transformer encoder in the FPDSC (turn-level). In the comparative network (single),  $\{c_1^s, \dots, c_{t-1}^{s}, m_t^{s,tl}\}$ are fed to the transformer encoder. Table \ref{single turn comparative result} shows the complete previous dialogue states improve the general performance of the model.

\begin{table}
\small
\begin{tabular}{c|c|c|c|c}
\hline
Dataset & single$^{*}$ & single & turn-level$^{*}$ & turn-level \\
\hline
dev  & 57.25 & 60.94 & 56.17 & 61.39 \\
\hline
test &  54.40 & 56.70 & 54.08 & 57.88 \\
\hline
\end{tabular}
\caption{Joint accuracy results of the comparative network (single) and the FPDSC (turn-level) on the MultiWOZ 2.1 dataset. $^{*}$ means that the approach is only trained in the teacher-forcing training, otherwise is further trained by uniform scheduled sampling training after the teacher-forcing phase.}
\label{single turn comparative result}
\end{table}

\subsection{Deleted-value Tests}
We select dialogues containing the deleted-value problem from test data in MultiWOZ 2.1. We regard the above dialogues as templates and augment the test data by replacing the original slot value with other slot values in the ontology. There are 800 dialogues in the augmented data. We only count the slots in dialogue turn,  which occurs the deleted-value problem.  As shown in Table \ref{MUL2359}, if \textit{restaurant-name=rice house} at the $6^{th}$ turn and \textit{restaurant-name=None} at the $7^{th}$ turn, we regard it as a successful tracking. We use the success change rate to evaluate the effectiveness. Table \ref{value_delete_related} shows that the explicit introduction of the previous dialogue states in both turn-level and passage-level helps solve the problem.

\begin{table}
\small
\begin{tabular}{l}
\hline
$U_{1}$:Find me a museum please\\
\textbf{restaurant-name: None}\\
\hline
$S_{2}$:There are 23 museums. Do you have an area as \\preference?\\
$U_{2}$:I just need the area and address for one of them.\\
\textbf{restaurant-name: None}\\
\hline
$S_{3}$:I have the broughton house gallery in the centre \\at 98 king street.\\ 
$U_{3}$:Thank you so much. I also need a place to dine \\in the centre that serves chinese food.\\
\textbf{restaurant-name: None}\\
\hline
$S_{4}$:I have 10 place in the centre. Did you have a price \\range you were looking at? \\
$U_{4}$:I would like the cheap price range.\\
\textbf{restaurant-name: None}\\
\hline
$S_{5}$:I recommend the rice house. Would you like me \\to reserve a table?\\
$U_{5}$:yes, please book me a table for 9 on monday at 19:30.\\
\textbf{restaurant-name: rice house}\\
\hline
$S_{6}$:Unfortunately, I could not book the rice house for \\that day and time. Is there another day or time that would \\work for you?\\
$U_{6}$:Can you try a half hour earlier or later and see if the \\have anything available?\\
\textbf{restaurant-name: rice house}\\
\textbf{Dual-level: restaurant-name: rice house}\\
\textbf{Base: restaurant-name: rice house}\\
\hline
$S_{7}$:No luck, would you like me to try something else?\\
$U_{7}$:Yes, please find another cheep restaurant for that \\amount of people at that time.\\
\textbf{restaurant-name: None}\\
\textbf{Dual-level: restaurant-name: None}\\
\textbf{Base: restaurant-name: rice house}\\
\hline
\end{tabular}
\caption{\label{MUL2359} Dialogue id MUL2359 from MultiWOZ 2.1}
\end{table}

\subsection{Related-slot Tests}
We focus on the multi-domain dialogues which contain dialogue domain of taxi for the related-slot test. We select 136 dialogue turns from the MultiWOZ 2.1 test data, which contains the template such as \textit{book a taxi from A to B} or \textit{commute between A and B}. We replace the explicit expression in order to focus on the actual related-slot filling situation. For example, in the dialogue from Table \ref{MUL2657}, we replace the value \textit{Ballare} to \textit{attraction} in the user utterance at the $4^{th}$ turn. We only count slots \textit{taxi-departure} and \textit{taxi-destination} without value of \textit{None} in the dialogue turns, which contain the related-slot phenomenon. We divide the sum of successful tracking counts by the number of the above slots to get the success change rate. Table \ref{value_delete_related} shows the result.

\begin{table}
\small
\begin{tabular}{l}
\hline
$U_{1}$: Can you give me information on an attraction \\called ballare?\\
\textbf{taxi-departure: None;taxi-destination: None}\\
\hline
$S_{2}$: The Ballare is located in Heidelberg Gardens, \\Lion Yard postcode cb23na, phone number is \\01223364222. The entrance fee is 5 pounds.\\
$U_{2}$: Thanks. I'm also looking for somewhere to stay\\ in the north. It should be in the moderate price range \\and has a star of 2 as well\\
\textbf{taxi-departure: None;taxi-destination: None}\\
\hline
$S_{3}$: Would you want to try the lovell lodge, \\which is in the moderate price range and in the north.\\
$U_{3}$: Let's do that. Please reserve it for 6 people and\\ 5 nights starting from thursday.\\
\textbf{taxi-departure: None;taxi-destination: None}\\\hline
$S_{4}$: The booking goes through and the reference \\number is TY5HFLY1.\\
$U_{4}$: Can you help me to book a taxi from the \\hotel to the Ballare. I want to leave by 17:30.\\
\textbf{taxi-departure: lovell lodge}\\
\textbf{taxi-destination: ballare};\textbf{taxi-leave: 17:30}\\
\hline
\end{tabular}
\caption{\label{MUL2657} Dialogue id MUL2657 from MultiWOZ 2.1}
\end{table}

\begin{table}
\small
\begin{tabular}{c|c|c|c|c}
\hline
Joint Acc &\multicolumn{2}{c}{Normal Evaluation} &\multicolumn{2}{|c}{Evaluation with}\\
$\%$&\multicolumn{2}{c}{}&\multicolumn{2}{|c}{Teacher Forcing}\\
\hline
Dataset& dev & test & dev & test\\
\hline
Base & 58.01 &54.91 & $-$ & $-$\\
\hline
Turn-level$^{*}$ & 56.17 & 54.08 & 69.13 & 65.82\\
\hline
Turn-level & 61.39 & 57.88 & $-$ & $-$ \\
\hline
Passage-level$^{*}$ & 55.21 & 52.40 & 66.84 & 61.92\\
\hline
Passage-level & 61.11 & 55.86 & $-$ & $-$\\
\hline
Dual-level$^{*}$ & 56.17 & 54.08 & 70.22 & 67.17 \\
\hline
Dual-level & 61.89 & 59.07 & $-$ & $-$\\
\hline
\end{tabular}
\caption{\label{phase_study} Joint accuracy results of variants of our approach in different training phase on MultiWOZ 2.1. Normal evaluation means that the approach uses predicted dialogue states as inputs. Evaluation with teacher forcing means that it uses truth label as previous dialogue states. $^{*}$ means that the approach is only trained in teacher-forcing training, otherwise is further trained by uniform scheduled sampling training after the teacher-forcing phase.}
\end{table}

\begin{figure*}[t]
\centering
\resizebox{1\linewidth}{!}{
\includegraphics[width=25cm]{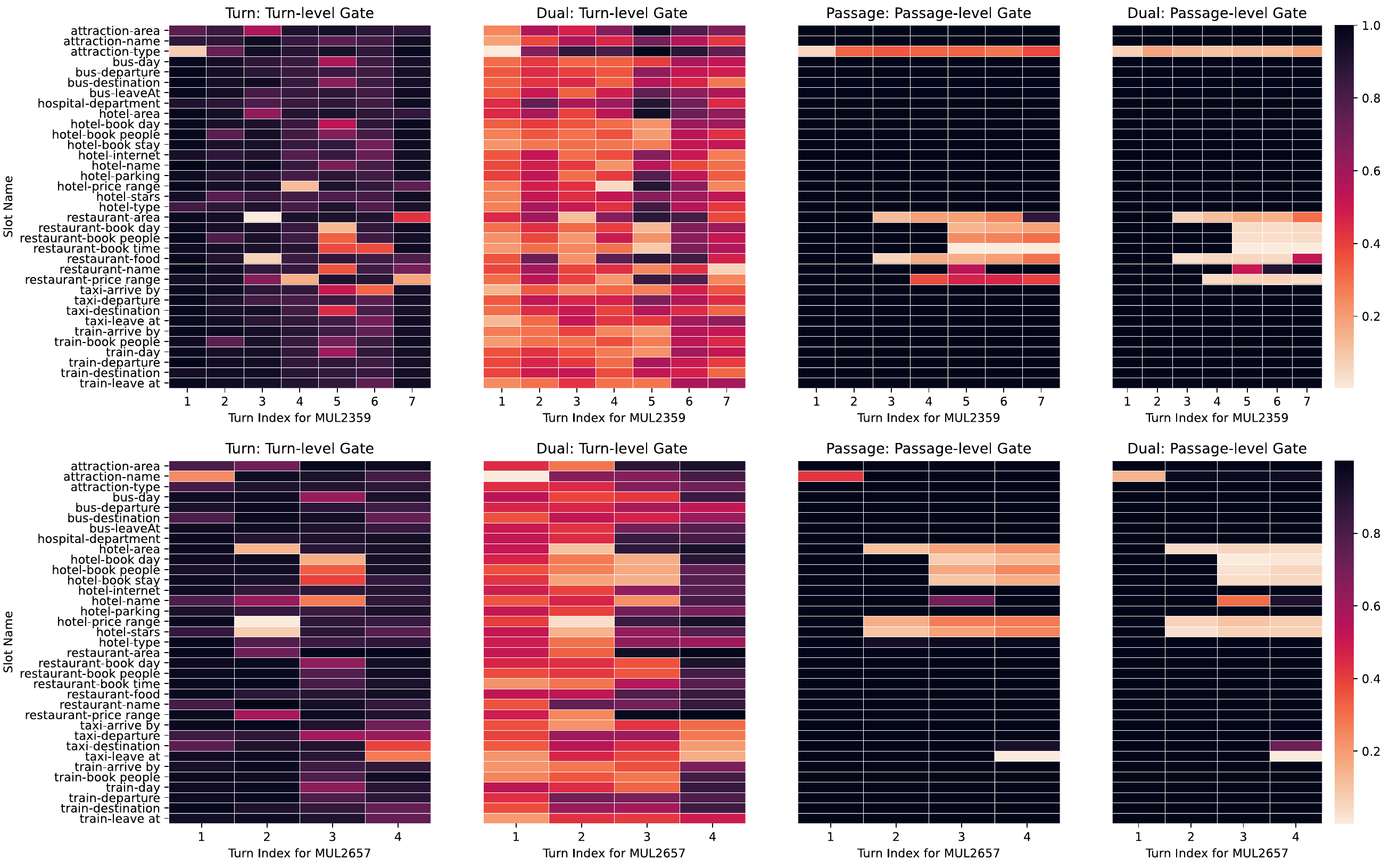}}
\caption{Visualization of output weights in the fusion gates. The weight represents the proportion of the information from the previous dialogue states. The large weight with dark color means that the prediction of the slot value pays much attention to the previous dialogue states. Turn, Passage, Dual mean FPDSC with turn-level, passage-level and dual-level.}
\label{slot gate}
\end{figure*}

\subsection{Gate Visualization}
Figure \ref{slot gate} shows the output weight of the turn/passage-level fusion gates in dialogue \textbf{MUL2359} (Table \ref{MUL2359}) and \textbf{MUL2657} (Table \ref{MUL2657}) from MultiWOZ 2.1. \textbf{Turn}, \textbf{Passage}, \textbf{Dual} in titles of subplots represent FPDSC with turn-level, passage-level, and dual-level. All the weights in Figure \ref{slot gate} mean the information keep-proportion from the last dialogue states. 

When we focus on the slot \textbf{restaurant-name} in dialogue \textbf{MUL2359}. The output weight in the turn-level fusion gate is small at the $5^{th}$ and the $7^{th}$ dialogue turn in turn/dual-level approaches. Since the slot value \textbf{rice house} is first mentioned at the $5^{th}$ turn and the constraint is released at the $7^{th}$ turn, the change of the weight for slot \textbf{restaurant-name} is reasonable. When we focus on slots \textbf{taxi-departure}, \textbf{taxi-destination}, and \textbf{taxi-leave at} at the $4^{th}$ turn of dialogue \textbf{MUL2657}, the respective information sources for above three slots are only previous dialogue state (\textbf{hotel-name} to \textbf{taxi-departure}), both previous dialogue state and current user utterance (\textbf{Ballare} can be found in both user utterance and previous dialogue states of \textbf{attraction-name}), only user utterance (\textbf{17:30} appears only in the user utterance at the $4^{th}$ dialogue turn). As shown in Figure \ref{slot gate}, at the $4^{th}$ dialogue turn of \textbf{MUL2657}, \textbf{taxi-departure} has a large weight, \textbf{taxi-destination} has a middle weight, \textbf{taxi-leave at} has a small weight. This trend is as expected.

Figure \ref{slot gate} also shows that the turn-level fusion gate is sensitive to signal tokens in the current user expression. At the 4$^{th}$ dialogue turn of \textbf{MUL2359}, the word \textbf{cheap} triggers low output weight of the turn-level fusion gate for slots \textbf{hotel-price range} and \textbf{restaurant-price range}. It is reasonable that no domain signal is in the 4$^{th}$ utterance. The output of the passage-level fusion gate will keep a relatively low weight once the corresponding slot is mentioned in the dialogue except for the name-related slot.

Although the output weights of the passage-level fusion gate share similar distribution in passage/dual-level method at the $7^{th}$ dialogue turn of \textbf{MUL2359}. FPDSC (passage-level) has a false prediction of \textbf{restaurant-name} and FPDSC (dual-level) is correct. Two fusion gates can work together to improve the performance. It explains the high performance in dual-level strategy.

\subsection{Ablation Study}
Table \ref{Joint Acc} shows that the passage/turn/dual-level approaches get improvements ($0.95\%$, $2.97\%$, $4.16\%$) compared to the base approach in MultiWOZ 2.1. The results show the turn-level fusion gate is vital to our approaches. The entire history of dialogue states is helpful for DST.
The uniform scheduled sampling training is crucial for improving our models' performance. In Table \ref{phase_study}, \textbf{dev} and \textbf{test} represent validation and test data. As the table shows, all of our approaches improve the joint accuracy around $3\%$ after uniform scheduled sampling training. The falsely predicted dialogue states work as the data noise, which improves the model's robustness. The base approach utilizes only the information from the context without uniform scheduled sampling training. 

\section{Conclusion}
In this paper, we combine the entire history of the predicted dialogue state and the contextual representation of dialogue for DST. We use a hierarchical fusion network to merge the turn/passage-level information. Both levels of information is useful to solve the deleted-value and related-slot problem. Therefore, our models reach state-of-the-art performance on MultiWOZ 2.0 and MultiWOZ 2.1.

The turn-level fusion gate is sensitive to signal tokens from the current turn utterance. The passage-level fusion gate is relatively stable. Uniform scheduled sampling training is crucial for improving our models' performance. The entire history of dialogue states helps at extracting information in each dialogue utterance. Although some errors exist in the predicted dialogue states, the errors work as the data noise in the training to enhance the proposed model's robustness.

Although our approach is based on predefined ontology, the strategy for information extraction is universal. Besides, the core feature $f_t^{s}$ can be introduced to a decoder to generate the slot state, which suits most open-domain DST.

\section*{Acknowledgement}
We thank the anonymous reviewers for their helpful comments. This work is supported by the NSFC projects (No. 62072399, No. 61402403), Hithink RoyalFlush Information Network Co., Ltd, Hithink RoyalFlush AI Research Institute, Chinese Knowledge Center for Engineering Sciences and Technology, MoE Engineering Research Center of Digital Library, and the Fundamental Research Funds for the Central Universities. 
 
\bibliographystyle{acl_natbib}
\bibliography{acl2021}


\end{document}